\documentclass[3p,review,12pt]{elsarticle}

\usepackage{hyperref,url}
\usepackage{lineno}
\usepackage{booktabs}
\usepackage{multirow}
\usepackage{tcolorbox}
\usepackage{amsmath}
\usepackage{amsfonts}
\usepackage{stfloats}
\usepackage{adjustbox}
\usepackage{algorithmic}
\usepackage{algorithm}
\usepackage{booktabs}
\usepackage{multirow}
\usepackage{mathtools}
\modulolinenumbers[5]
\begin{document}

\begin{frontmatter}

\title{Language Modeling for Code-Switched Data: Challenges and Approaches}

\author{Ganji Sreeram\corref{mycorrespondingauthor}}
\ead{s.ganji@iitg.ernet.in}
\author{Rohit Sinha}
\ead{rsinha@iitg.ernet.in}
\address{Department of Electronics and Electrical Engineering, \\Indian Institute of Technology Guwahati, Guwahati-781039, India.}
\cortext[mycorrespondingauthor]{Corresponding author}

\begin{abstract}
Lately, the problem of code-switching has gained a lot of attention and has emerged as an active area of research. In bilingual communities, the speakers commonly embed the words and phrases of a non-native language into the syntax of a native language in their day-to-day communications. The code-switching is a global phenomenon among multilingual communities, still very limited acoustic and linguistic resources are available as yet. For developing effective speech based applications, the ability of the existing language technologies to deal with the code-switched data can not be over emphasized. The code-switching is broadly classified into two modes: inter-sentential and intra-sentential code-switching. In this work, we have studied the intra-sentential problem in the context of code-switching language modeling task. The salient contributions of this paper includes: (i) the creation of Hindi-English code-switching text corpus by crawling a few blogging sites educating about the usage of the Internet (ii) the exploration of the parts-of-speech features towards more effective modeling of Hindi-English code-switched data by the monolingual language model (LM) trained on native (Hindi) language data, and (iii) the proposal of a novel textual factor referred to as the code-switch factor (CS-factor), which allows the LM to predict the code-switching instances. In the context of recognition of the code-switching data,  the substantial reduction in the PPL is achieved with the use of POS factors and also the proposed CS-factor provides independent as well as additive gain in the PPL.  
\end{abstract}
\end{frontmatter}

\section{Introduction}
\label{sec:introduction}
Code-switching is defined as the alternate use of two or more languages by a speaker within the same utterance during conversation~\cite{Gumperz_1982_Discourse}. Over the years, due to urbanization and geographical distribution, people have moved from one place to another for a better livelihood. Communicating in two or more languages helps to interact better with people from different places and cultures. Code-switching differs from other types of language mixing such as borrowing, pidgin, etc. Code-switching refers to the phenomena where the bilingual speaker switches between the two languages while conversing with another person who also have the knowledge about both the languages. On the other hand, borrowing is referred as embedding words from the foreign language into the native language due to the absence of those words in the vocabulary of the native language~\cite{Myers_1992_Comparing}. Pidgin is a grammatically simplified means of communication that develops between two or more groups that do not have a language in common. After independence, though the Indian constitution declared Hindi as the primary official language, the usage of English was still continued as secondary language for its dominance in administration, education and law~\cite{dey2014hindi, malhotra1980hindi}. Thereby, creating a trend among the urban population to communicate in English for economic and social purposes. Over the years, substantial code-switching to English while speaking Hindi as well as other dominant Indian languages has become a common feature. Despite extensive code-switching, the speakers and the listeners mostly agree upon which language the code-switched sentence mainly belongs to. That particular language is referred to as native (matrix) language and the other is the non-native (embedded) language~\cite{Joshi_1982_Processing}. The people belonging to the bilingual communities say that the main reason for code-switching between languages is due to the lack of words in the vocabulary of that particular native language~\cite{Grosjean_1982_Life}. Other possible reasons for code-switching are: (i) to qualify the message by emphasizing specific words, (ii) to convey a personalized message, (iii) to maintain confidentiality during verbal communication, (iv)  to show expertise, authority, status, etc.
 
\begin{figure}[]
\begin{minipage}[b]{1.0\linewidth}
  \centering
  \centerline{\includegraphics[width=14cm]{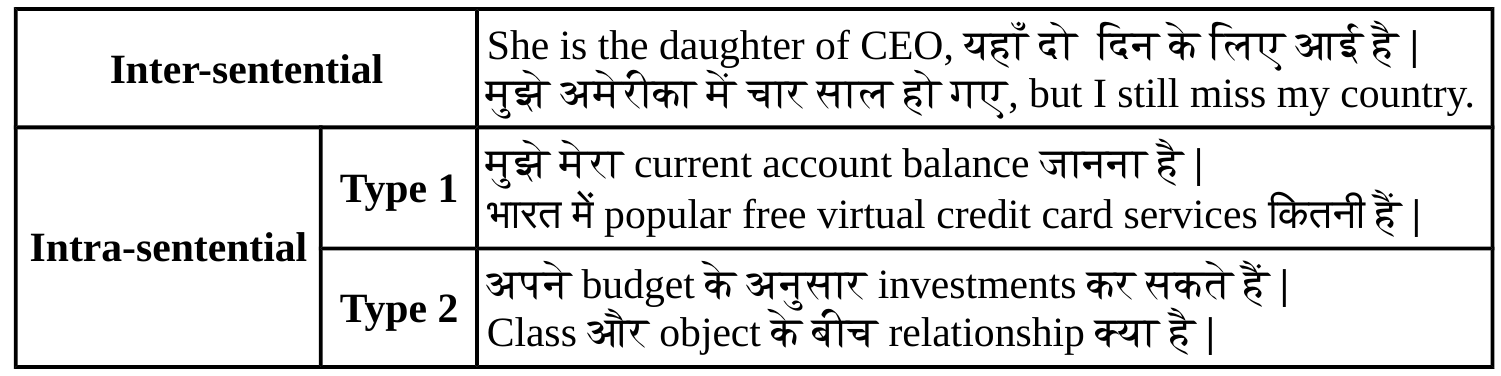}}\vspace{-3mm}
  \caption{Example sentences showing the inter-sentential code-switching and variants of intra-sentential code-switching. Type~1 and Type~2 refer to those cases where the embedded English words carry high and low contextual information, respectively. In this work, we mainly deal with Type~2 intra-sentential code-switching problem.}
  \label{cs_ex}
\end{minipage}
\end{figure}

The switching of languages can happen within the sentence or at its boundary. The switching within the sentence is referred to as the intra-sentential code-switching and the one happening at the sentence boundary is referred to as the inter-sentential code-switching~\cite{Grosjean_1982_Life}.  In this paper, we address the intra-sentential code-switching mode in Indian context. Here, we consider the code-switching between Hindi and English languages, where the native language is Hindi and the non-native (foreign) language is English~\cite{Bhuvanagirir_2012_Mixed}. In literature, there exists many works addressing the intra-sentential code-switching problem~\cite{ahmed2012automatic, Cao_2010_Semantics}.  For training the bilingual language model (LM), in~\cite{Gumperz_1982_Discourse} the authors merged the training texts with possible tagging of words, which have the same spelling in more than one of the considered languages. Thus, they distinguish these words and build a language model of the merged training texts. A few other works also employed the interpolated LM where the vocabulary of both the languages are merged, but estimate the LM probabilities for each individual language. Word entries of one language are included in the language models of other languages with zero counts and received a small backoff probability. In that approach, all the words have non-zero probabilities which enabled them to handle the intra-sentential code-switching. But in most of those works, the non-native language contained word sequences having contextual information~\cite{Lyu_2006_Speech, Yeh_2010_Integrated}. In this work, we consider a case where most of the native (Hindi) language sentences are embedded with non-native (English) language words without much context information. Figure~\ref{cs_ex} shows the  examples of inter-sentential code-switching and different types of intra-sentential code-switching sentences highlighting the difference in the contextual information carried by the switched foreign words. Due to lack of sufficient context information, the existing approaches become less effective in recognizing such code-switching data. To address these challenges, one way is to tediously collect large amount of text corpus in code-switching domain for training the models as attempted in \cite{Lyu_2008_Language}. Alternatively, we should explore augmenting the monolingual LM with code-switching information.

The remainder of this paper is organized as follows: In Section~\ref{sec:motivation}, the motivation of the work is presented followed by the discussion on the explored parts-of-speech (POS) and the proposed code-switching textual features in training recurrent neural network based language model (RNN-LM) in~\ref{sec:propose}. The description of the text corpora and the system parameters involved in this study are presented in Section~\ref{sec:setup}. The evaluation results of the proposed code-switch factor (CS-factor) based RNN-LM in contrast with the existing approaches has been presented in Section~\ref{sec:results}. The paper is concluded in Section~\ref{sec:conclusion}. 

\section{Motivation of the work}
\label{sec:motivation}  
 Intra-sentential code-switching is not a random mixing of one language with the other, rather the switching happens with the systematic interactions between the two languages. It is hypothesized that, if the semantic and syntactic information about the native words that are being code-switched with the foreign words can be captured, the same could enhance the ability of the monolingual LM to deal with code-switching task. Based on this intuition, we explore the parts-of-speech (POS) features from the native Hindi language text data. Factored language models (LMs) have been introduced to incorporate morphological information in language modeling~\cite{Kirchhoff_2007_Factored,Bilmes_2003_Factored,Schultz_Speech}. In the factored language model, each word $w_{t}$ in the vocabulary $V$ is represented as a collection of $k$ factors: $f_{t}^{1},\, f_{t}^{2},\, \ldots, \, f_{t}^{k}$. The factors of a word can be anything, including morphological classes, stems, roots, and any other linguistic features that might correspond to that word. A factor can even be a word itself, so that the probabilistic language modeling covers both words and their factors. Factored language models can be applied to any language which is semantically accepted since the semantic factors like the parts-of-speech can be used as factors. The recurrent neural networks (RNNs) possess the ability to model the long-term dependencies and also the semantics information more efficiently compared to the traditional $n$-gram scheme in context of LMs~\cite{mikolov2013efficient, backoff_oparin_2012, cued_rnnlm_chen_2015}. Recently, the RNNs are also employed in training the factored LMs and have shown significant gains in terms of the perplexity (PPL) and the error rate in speech recognition task~\cite{wu2012factored, adel2013recurrent}. Motivated by those facts, we have explored the parts-of-speech (POS) factors in training the native (Hindi) language LM using RNNs in this work. Later, the Hindi-English code-switching data is tested over that factored RNN-LM.   Following that, we introduce a novel textual factor that allows to include the code-switching information in training of the factored RNN-LMs.
 \begin{figure}[]
 \begin{minipage}[b]{1.0\linewidth}
   \centering
   \centerline{\includegraphics[width=12cm]{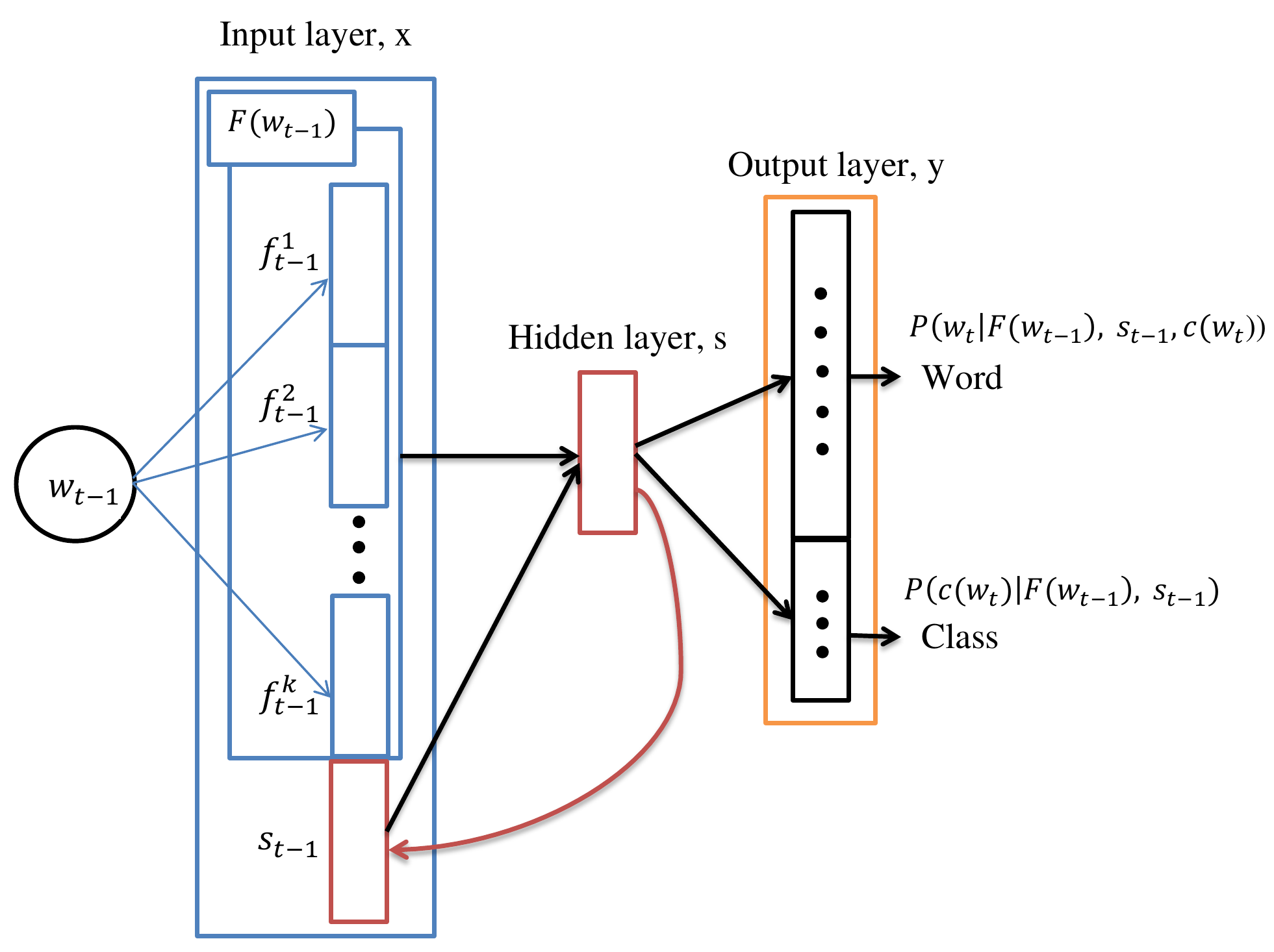}}\vspace{-2mm}
   \caption{Block  diagram of the factored RNN-LM.}
   \label{fig:frnnlm}
 \end{minipage}
 \end{figure}
\section{Proposed Approach}
\label{sec:propose}
In this work, we have explored the factored RNN-LM for code-switching task. For that purpose, the POS textual factors are extracted from Hindi data  and are used along with the words in modeling. The network architecture of the factored RNN-LM is given in Figure~\ref{fig:frnnlm}, where the input, hidden and the output layers are denoted by $x$, $s$ and $y$, respectively. The factored RNN-LM predicts the posterior probability of the current word as
\begin{equation}
P(w_{t}|F(w_{t-1}), s_{t-1})=P(w_{t}|F(w_{t-1}), s_{t-1},c(w_{t})) \times \,P(c(w_{t})|F(w_{t-1}), s_{t-1}) 
\end{equation}  \label{eq1}
where $F(w_{t-1})=\left[f_{(t-1)}^{1},\, f_{(t-1)}^{2},\, \ldots,\, f_{(t-1)}^{k}\right]$. 
\begin{figure}[]
\begin{minipage}[b]{1.0\linewidth}
  \centering
  \centerline{\includegraphics[width=12cm]{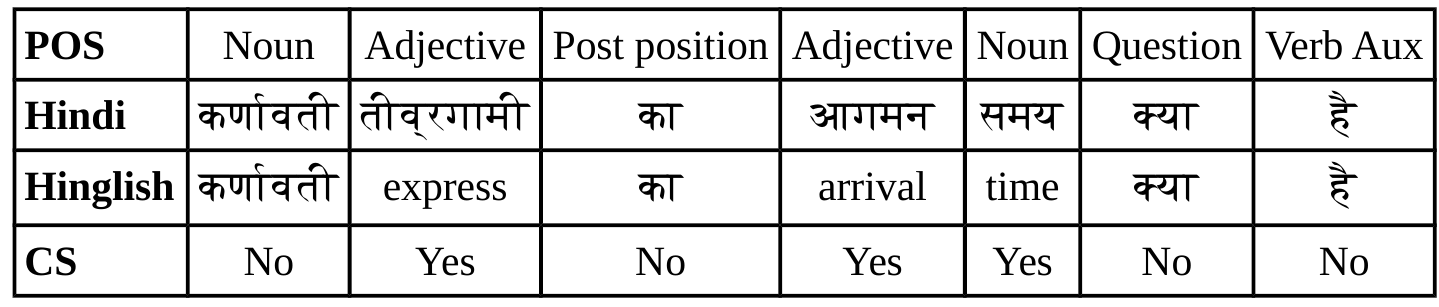}}\vspace{-3mm}
  \caption{An example pair of Hindi and Hindi-English (Hinglish) sentences and their factors. It is worth highlighting that the POS labels extracted for Hindi text remain mostly valid for Hinglish text too. By aligning Hinglish and Hindi sentences on word basis, the code-switch (CS) factors can be labeled.}
  \label{fig:ex}
\end{minipage}
\end{figure}

\subsection{Parts-of-Speech (POS) Factors}
In natural language processing, the parts-of-speech of the words are derived given the context information. The POS mainly depends on the adjacent words in a given sentence. The POS tagging is not just tagging a list of words with their respective parts-of-speech information. Based on the context, the same word might have more than one POS factors at different locations. When these POS factors are used as the features along with the words while training the LM, it is expected to model the contextual information in more effective way for the code-switched data. We hypothesize that, in majority of the code-switching cases only the native language words are replaced by the foreign language words. Whereas the context and the parts-of-speech of that replaced foreign language word will be mostly same as that of the native language word. This can be visualized from the example given in Figure~\ref{fig:ex}. Also, while training the factored RNN-LM, we have added all the foreign (English) language words from the code-switching training data to the native (Hindi) language text data to avoid the out-of-vocabulary (OOV) issue. However, no effort has been made to address the OOV  for the native (Hindi) language. The parts-of-speech tagging of the Hindi training data has been done using an existing Hindi POS tagger~\cite{pos_tag}.

 The POS tagger is a tool that helps achieve the part-of-speech labeling in a sentence. Most of the Indian languages do not have POS taggers due to lack of linguistic resources such as text corpora, morphological analyzers, lexicons, etc. Recent research has shown that if the low resourced language is typologically related to a rich resourced language, it is possible to  build taggers for the former using the linguistic resources of the later~\cite{reddy2011cross}. On studying the topological similarity between Hindi-English and their Hindi translated versions, we notice a good match. In our collected data, we find that about $70\%$ sentences involve only the replacement of English words with their Hindi counterparts without need of any change in the syntax of the sentences. Motivated by that the POS factors for the Hindi-English code-switched data are extracted using the POS tagger trained on the Hindi language database.  Wherever the said tool is not able to identify any part of the given sentence correctly, it outputs 'unknown' (UNK) label. 
 
\subsection{Code-Switch (CS) Factor}
During code-switching the foreign language words are inserted into the native language sentences mostly without altering the semantics and syntactics of the native language. Thus, like the POS, the code-switching is also expected to adhere certain semantic and syntactic rules. To capture this information, we have proposed a novel factor that identifies the locations where the code-switching can potentially occur or not. This factor is referred to as the CS-factor in this work. To address the OOV issue due to foreign (English) words, we have followed the same procedure which is discussed in the context of the POS factors. When a monolingual LM is trained by adding the CS-factor along with the words and tested using the code-switched data, a significant improvement in terms of the perplexity is noted over the LMs created with and without the POS factors. 

To introduce the CS-factor in the LM training data, a separate tagger is required but the same is yet to be developed. For quick validation of the idea, in this work, the CS-factors are derived by following a simple scheme. For doing that, we have aligned the Hindi-English training sentences with their Hindi translated versions, to capture the information about the positions where code-switching occurs. The words which undergo code-switching are marked as `Yes' while the remaining non-switched words are marked as `No' as shown in  Figure~\ref{fig:ex}. The Hindi training set is then tagged using the CS-factor based on the mapping learnt.  We note that the data collected for our experiments involve $70\%$ sentences where the English words are replaced with their Hindi counterparts without any change in the syntax. Hence, the CS factors for the Hindi-English code-switched data are kept the same as that of Hindi data while training the Hindi-English code-switched LMs.

\section{Experimental Setup}
\label{sec:setup}
 This section describes the text data used in the experimentation, the extraction of the POS and the proposed CS factors, and the parameter used for training the factored LMs. 
 \begin{figure}[]
 \begin{minipage}[b]{1.0\linewidth}
   \centering
   \centerline{\includegraphics[width=13.1cm]{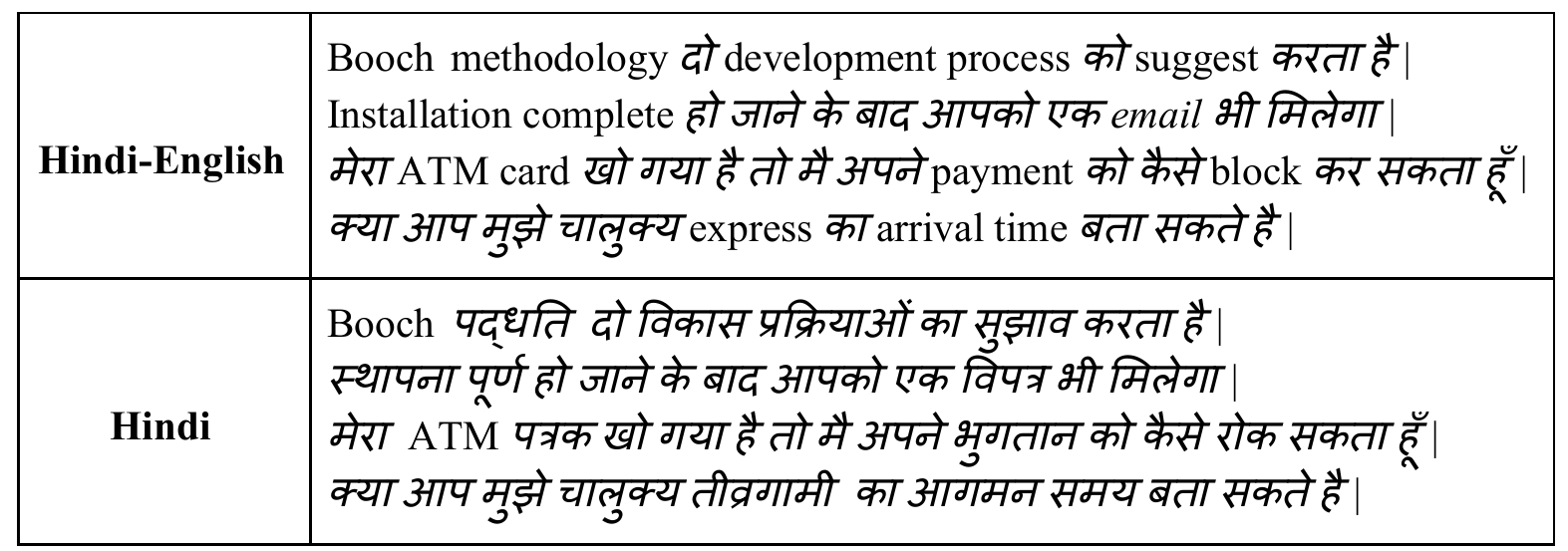}}
   \caption{Typical Hindi-English code-switched sentences in the database and their respective Hindi translated versions. It is to note that proper-nouns and abbreviations are retained in Hindi version.}
   \label{fig:ex_data}
 \end{minipage}
 \end{figure}
\subsection{Database preparation}
\label{tagging}
\begin{figure}[]
 \begin{minipage}[b]{1.0\linewidth}
   \centering
   \centerline{\includegraphics[width=16.5cm]{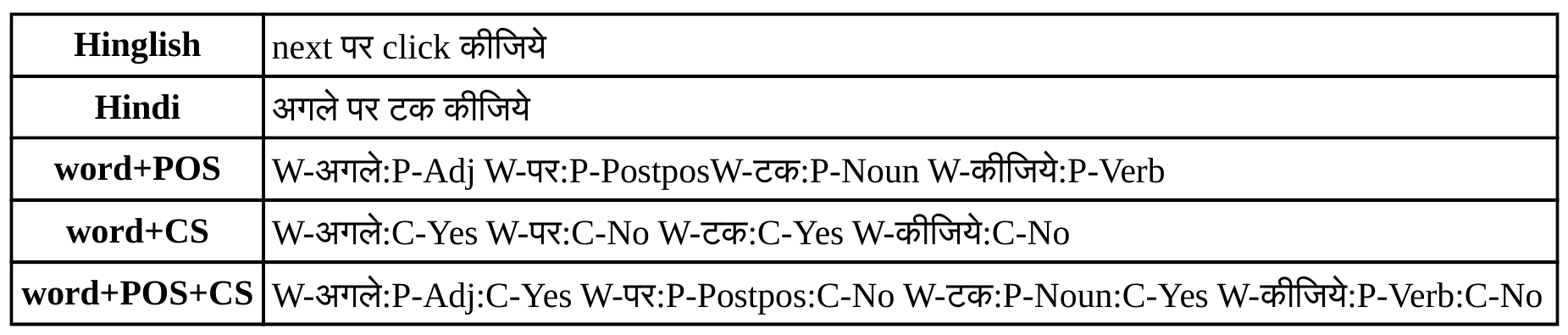}}
   \caption{The syntax of different transcriptions that are used for training the factor-based RNNLM. Here, P and C refers to the POS and the CS factors respectively.}
   \label{fig:ex_factors}
 \end{minipage}
 \end{figure}

For the experimental purpose, the code-switched Hindi-English text data has been collected by crawling a few blogging websites educating about the Internet usage\footnote{https://shoutmehindi.com}$^{,}$\footnote{https://notesinhinglish.blogspot.in}. The crawled text data has been normalized into meaningful sentences while removing the special characters, emoticons, spaces, etc. Later, these Hindi-English sentences are translated to Hindi sentences primarily with the help of {\it Google translate}\footnote{https://translate.google.com/} online translation tool. In some cases, the Google translate failed to produce correct Hindi translation possibly due to inadequate Hindi vocabulary. In such cases, the Hindi translation has been done manually with the help of a few online Hindi vocabulary resources~\footnote{http://www.rajbhasha.nic.in/hi/hindi-vocabulary}$^{,}$\footnote{https://hi.wiktionary.org/wiki}. It is to note that the proper-nouns and the abbreviations present in the Hindi-English sentences are kept unchanged while translating them to Hindi. A few example Hindi-English sentences and their Hindi translated versions are shown in Figure~\ref{fig:ex_data}. The obtained Hindi text data is divided into training and test sets consisting of $1050$ and $105$ sentences, respectively. Similarly, the Hindi-English text data is also partitioned into identical sized training and test sets while maintaining one-to-one correspondence with those of Hindi data sets. The salient details of different Hindi-English and Hindi data sets are summarized in Table~\ref{database}. Also, the syntax of various transcriptions that are used for training the factor-based RNNLMs is given in Figure~\ref{fig:ex_factors}. 
\begin{table}[]
\centering
\caption{Description of the data sets created for evaluating the code-switching task. The reason behind English words in Hindi data lies in the proper-nouns and abbreviations being left unchanged while translating Hindi-English (Hinglish) sentences to Hindi.}
\label{database}
\scalebox{0.77}{
\begin{tabular}{|c|c|c|c|c|c|c|c|c|c|c|}
\hline
\multirow{3}{*}{\textbf{Data}} & \multicolumn{5}{c|}{\textbf{Train Set}}                                                                                                                         & \multicolumn{5}{c|}{\textbf{Test Set}}                                                                                                                          \\ \cline{2-11} 
                      & \multirow{2}{*}{\begin{tabular}[c]{@{}c@{}}No. of \\ Sentences\end{tabular}} & \multicolumn{2}{c|}{No. of Words} & \multicolumn{2}{c|}{No. of Unique Words} & \multirow{2}{*}{\begin{tabular}[c]{@{}c@{}}No. of\\  Sentences\end{tabular}} & \multicolumn{2}{c|}{No. of Words} & \multicolumn{2}{c|}{No. of Unique Words} \\ \cline{3-6} \cline{8-11} 
                      &               & Hindi          & English          & Hindi         & English         &         & Hindi         & English        & Hindi            & English         \\ \hline \hline
Hinglish              & 1050          & 10484          & 5936             & 916           & 1418            & 105                		      & 1088          & 533            & 284              & 263             \\ \hline
Hindi                 & 1050          & 15604          & 1036             & 1938          & 381             & 105     		      & 1563          & 59             & 508              & 29              \\ \hline
\end{tabular}}
\end{table} 

The Hindi training data is used for developing the proposed and the contrast language models for the code-switching task. The Hindi test data is used for tuning the parameters while training the RNN-LMs. The recognition performances of the trained LMs are evaluated on the Hindi-English test data. Further, another set of class-based and POS-factor based RNN-LMs are trained on the Hindi-English training data and are used to benchmark the performance on the proposed approach. Also, the performances for the $5$-gram LMs trained on Hindi and Hindi-English data using SRILM toolkit~\cite{stolcke2002srilm} are computed for the reference purpose.

\subsection{Parameter tuning}
\begin{figure}[]
\centering
\includegraphics[width=7.5cm]{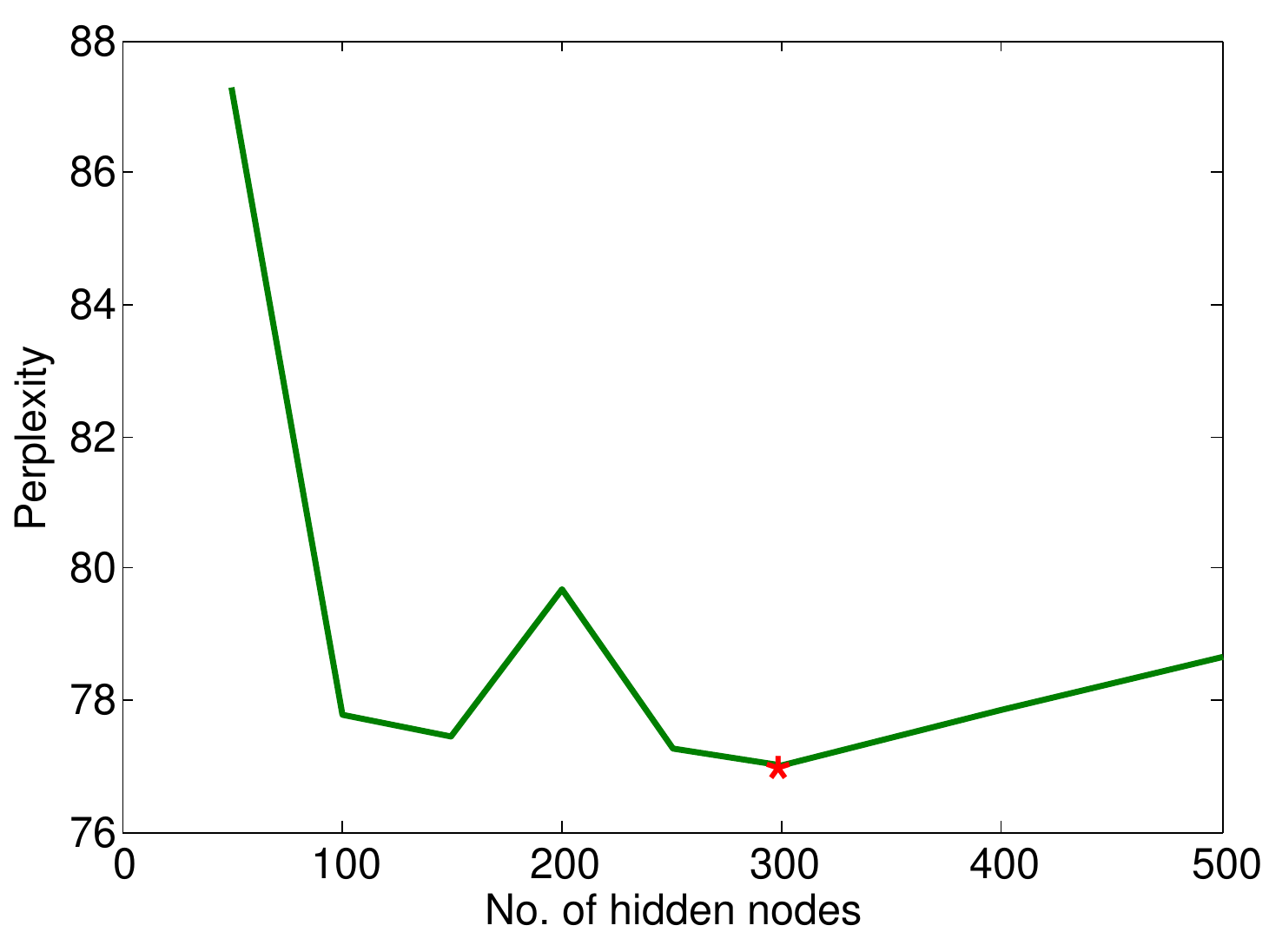}
\includegraphics[width=7.5cm]{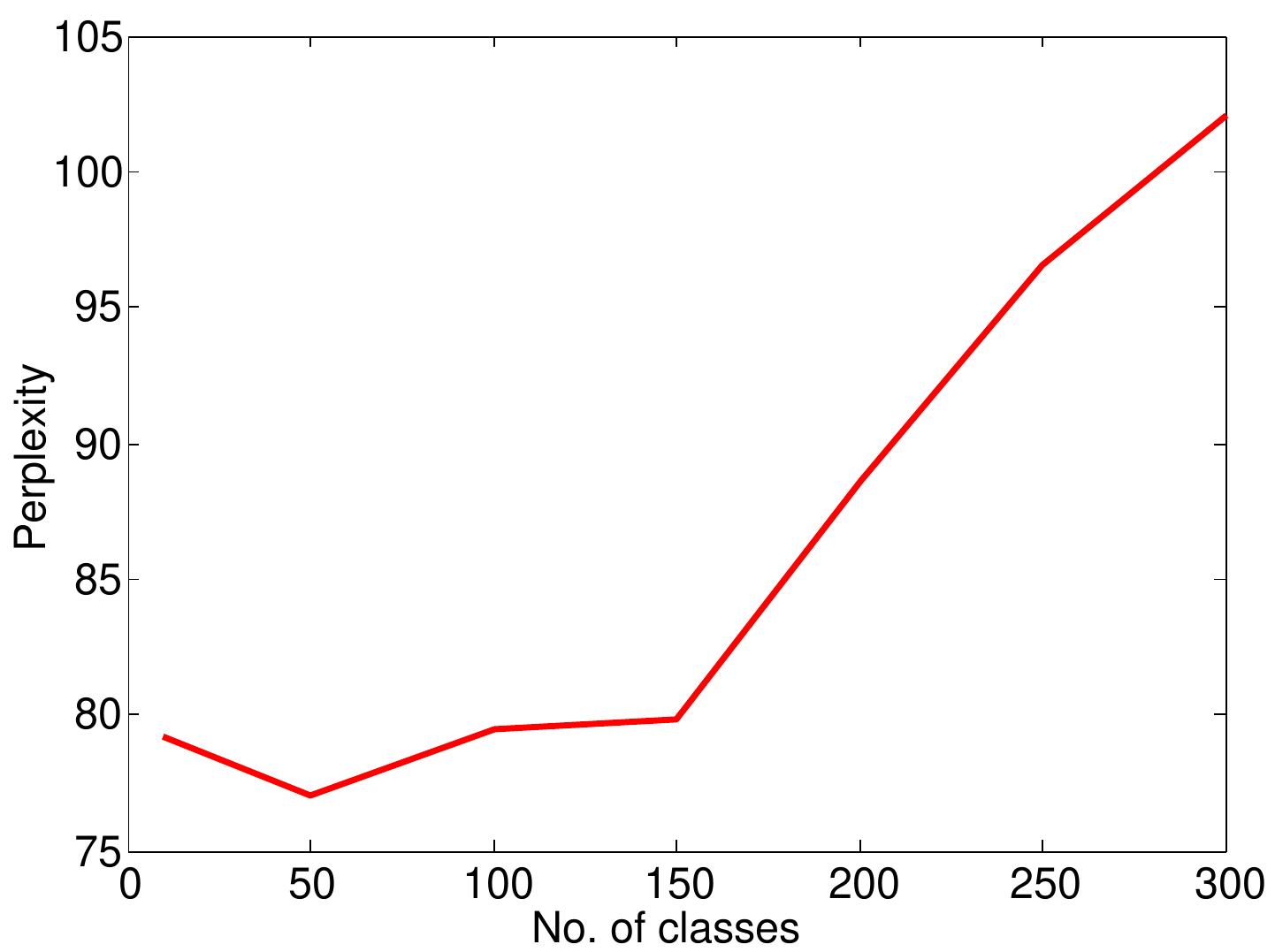}
  \caption{Tuning experiments for fixing the number of nodes in the hidden layer and the number of classes in training the RNN-LMs. First, the number of nodes is tuned using default settings of the toolkit used. Later, using the optimal value of the number of nodes as $300$, the number of classes is tuned.}
  \label{tune}
\end{figure}

RNN-based language models used in the experimental evaluation are developed using the RNNLM toolkit~\cite{mikolov_rnnlm_toolkit}. These LMs are trained with a single hidden layer having $300$ nodes with \emph{sigmoid} as the non-linearity function. By conducting tuning experiments on Hindi test data, the number of classes are set as $50$ and the variable corresponding to BPTT is set as $5$. Note that, the tuning of the number of classes selection has been done by considering the optimal value of the number of nodes in hidden layer. The perplexity values for different number of nodes in the hidden layer and for different number of classes is shown in Figure~\ref{tune}.  
\section{Results and Discussion}
\label{sec:results}
\begin{table}[]
\centering
\caption{The performances (in terms of perplexity) of the POS and the proposed CS-factors in training the Hindi RNN-LM in the context of Hindi-English (Hinglish) code-switching task. The performances of Higlish RNN-LMs are given for benchmarking. Since, the Hindi RNN-LMs being tuned on Hindi test set, those performances are for reference purpose only. Note that, the factored RNNLM is referred to as F-RNNLM.}\vspace{2mm}
\label{results}
\scalebox{1.18}{
\begin{tabular}{|c|c|c|c|r|}
\hline
\textbf{LM}                       & \textbf{Factor}                  & \textbf{Training data} & \textbf{Test data} & \multicolumn{1}{c|}{\textbf{PPL}} \\ \hline \hline
\multirow{3}{*}{\textbf{5-gram}}  & \multirow{3}{*}{word}            & Hinglish               & Hinglish           & 92.11                             \\ \cline{3-5} 
                                  &                                  & \multirow{2}{*}{Hindi} & Hinglish           & 334.20                            \\ \cline{4-5} 
                                  &                                  &                        & Hindi              & 80.11                             \\ \hline \hline
\multirow{3}{*}{\textbf{RNNLM}}   & \multirow{3}{*}{word}            & Hinglish               & Hinglish           & 89.67                             \\ \cline{3-5} 
                                  &                                  & \multirow{2}{*}{Hindi} & Hinglish           & 318.23                            \\ \cline{4-5} 
                                  &                                  &                        & Hindi              & 78.23                             \\ \hline \hline
\multirow{9}{*}{\textbf{F-RNNLM}} & \multirow{3}{*}{word + POS}      & Hinglish               & Hinglish           & 39.03                             \\ \cline{3-5} 
                                  &                                  & \multirow{2}{*}{Hindi} & Hinglish           & 58.12                             \\ \cline{4-5} 
                                  &                                  &                        & Hindi              & 58.91                             \\ \cline{2-5} 
                                  & \multirow{3}{*}{word + CS}       & Hinglish               & Hinglish           & 36.52                             \\ \cline{3-5} 
                                  &                                  & \multirow{2}{*}{Hindi} & Hinglish           & 51.67                             \\ \cline{4-5} 
                                  &                                  &                        & Hindi              & \multicolumn{1}{c|}{52.85}        \\ \cline{2-5} 
                                  & \multirow{3}{*}{word + POS + CS} & Hinglish               & Hinglish           & \multicolumn{1}{c|}{22.97}        \\ \cline{3-5} 
                                  &                                  & \multirow{2}{*}{Hindi} & Hinglish           & \multicolumn{1}{c|}{38.89}        \\ \cline{4-5} 
                                  &                                  &                        & Hindi              & \multicolumn{1}{c|}{39.55}        \\ \hline
\end{tabular}}
\end{table}
The evaluation of both the POS and the proposed CS factors have been done in the context of Hindi-English code-switching task. For this purpose, $5$-gram LMs and both the class-based and the factor-based LMs are developed separately using Hindi and Hindi-English training data. The performances of these systems in terms of perplexity (PPL) on Hindi-English and Hindi test data sets are reported in Table~\ref{results}. To avoid the paucity of data affect the findings, the $3$-fold cross-validation is done and the average performance of all those test sets has been reported. 

When the Hindi-English data is tested over the $5$-gram LM or the normal class-based Hindi RNN-LM, a huge degradation in PPL has been observed in comparison to that of the Hindi data. This is attributed to the fact that whenever a English words occur during testing, there is no context information with them. Note that, the OOV issue is taken care by adding all the English words in to the vocabulary of Hindi LM training data. Whereas, when the Hindi-English data is tested over factored Hindi RNN-LM trained using the POS information as a factor, a significant reduction in PPL is achieved. This is because, the POS factor is not only tagging a list of words with their respective parts-of-speech, but also is based on the context. Even when the native word is replaced by the foreign word, the POS factor will mostly remain unchanged. Thus, by employing the POS factors as a feature along with the words while training the RNN-LM, the context information will help predict the Hindi-Englihs word sequences. The parameters of all kinds of LMs are trained on Hindi test set, therefore the Hindi test set performances given in Table~\ref{results} are for reference purpose only. Further, from Table~\ref{results}, it can be noted that the PPL values of the Hindi-English test set on all factored RNN-LMs have turned out to be slightly better than those of the Hindi test set. This is attributed to the inclusion of English words towards addressing the OOVs in all types of Hindi RNN-LMs. As a result, the Hindi test set has higher OOV than the Hindi-English test set. From Table~\ref{results} we can see that the proposed CS-factor also resulted in significant improvement in the system performance over the conventional RNN-LM and also over the explored POS-based factored RNN-LM. Later, when the RNN-LM is trained by combining CS-factor along with the POS factors, further improvement in PPL is achieved. This result shows that the information captured by the CS-factor is additive to that of the POS factors.

For contrast purpose, Table~\ref{results} also lists the performances of the class-based RNN-LM created using Hindi-English training data when evaluated on the Hindi-English test data. Since the Hindi-English training data carries the Hindi language syntax for deriving the POS tags, it is  processed with the same Hindi POS tagger~\cite{reddy2011cross}. The obtained POS factors are then used to create the Hindi-English factored RNN-LM.  Also for comparison purpose, the recognition performances over the $5$-gram LMs trained using Hindi-English and Hindi text data are also reported. Interestingly, the proposed CS-factor inclusive factored monolingual RNN-LM has resulted in PPL value of $38.89$. 
\section{Conclusion}
\label{sec:conclusion}
In this work, we have explored the factored language model in the context of code-switching task. The factored native (Hindi)  language RNN-LM when created using both POS and proposed code-switching factors has shown significant reduction in perplexity for code-switched Hinglish test data. The proposed code-switching factor is simple to estimate yet found to be quite effective in handling the code-switched data. There is very little data available for the task considered in this work. In our experimentation, the collected data is still quite small for training the language models. So the validation on a bigger datasets is warranted and the same will be pursued in the future work.   
\bibliography{refs.bib}

\vspace{0.5cm}

\end{document}